\documentclass[runningheads]{llncs}

\usepackage{eccv}
\usepackage{eccvabbrv}
\usepackage{graphicx}
\usepackage{booktabs}
\usepackage{multirow}
\usepackage[accsupp]{axessibility}
\usepackage{hyperref}
\usepackage{orcidlink}
\usepackage{wrapfig}

\begin{document}

\title{AGCD: Agent-Guided Cross-modal Decoding for Weather Forecasting}
\titlerunning{AGCD for Weather Forecasting}

\author{
Jing Wu\inst{1}$^{*}$ \and
Yang Liu\inst{2}$^{*}$ \and
Lin Zhang\inst{3}$^{*}$ \and
Junbo Zeng\inst{1} \and
Jiabin Wang\inst{1} \and
Zi Ye\inst{1} \and
Guowen Li\inst{1} \and
Shilei Cao\inst{1} \and
Jiashun Cheng\inst{2} \and
Fang Wang\inst{4} \and
Meng Jin\inst{5} \and
YeRong Feng\inst{6} \and
Hong Cheng\inst{2} \and
Yutong Lu\inst{1,7} \and
Haohuan Fu\inst{8,7} \and
Juepeng Zheng\inst{1,7}$^{\dagger}$
}

\authorrunning{J. Wu et al.}

\institute{
Sun Yat-sen University, Zhuhai, China \and
The Chinese University of Hong Kong, Hong Kong, China \and
Jiangxi Science and Technology Normal University, Nanchang, China \and
China Meteorological Administration, Beijing, China \and
Huawei Technologies Co., Ltd, China \and
Guangdong-Hong Kong-Macao Greater Bay Area Weather Research Center for Monitoring Warning and Forecasting, China \and
National Supercomputing Center in Shenzhen, Shenzhen, China \and
Tsinghua University, Shenzhen, China
}

\maketitle

\begin{center}
\small
$^{*}$ Equal contribution \qquad $^{\dagger}$ Corresponding author
\end{center}

\begin{abstract}
Accurate weather forecasting is more than grid-wise regression: it must preserve coherent synoptic structures and physical consistency of meteorological fields, especially under autoregressive rollouts where small one-step errors can amplify into structural bias. 
Existing physics-priors approaches typically impose global, once-for-all constraints via architectures, regularization, or NWP coupling, offering limited state-adaptive and sample-specific controllability at deployment.
To bridge this gap, we propose Agent-Guided Cross-modal Decoding (AGCD), a plug-and-play decoding-time prior-injection paradigm that derives state-conditioned physics-priors from the current multivariate atmosphere and injects them into forecasters in a controllable and reusable way.
Specifically, We design a multi-agent meteorological narration pipeline to generate state-conditioned physics-priors, utilizing MLLMs to extract various meteorological elements effectively. To effectively apply the priors, AGCD further introduce cross-modal region interaction decoding that performs region-aware multi-scale tokenization and efficient physics-priors injection to refine visual features without changing the backbone interface.
Experiments on WeatherBench demonstrate consistent gains for 6-hour forecasting across two resolutions ($5.625^\circ$ and $1.40625^\circ$) and diverse backbones (generic and weather-specialized), including strictly causal 48-hour autoregressive rollouts that reduce early-stage error accumulation and improve long-horizon stability.
\keywords{Weather forecasting \and Multi-agent generation \and Physics-priors injection}
\end{abstract}

\section{Introduction}
Short-range weather forecasting is a cornerstone of operational prediction, underpinning public safety and high-stakes decision-making. High-impact phenomena can develop within hours, requiring accurate forecasts of evolving multi-variable atmospheric states that preserve cross-variable physical consistency. In this regime, small single-step errors, seemingly minor under grid-wise metrics, can accumulate and amplify into structural biases during autoregressive deployment. Traditionally, Numerical Weather Prediction (NWP) maintains consistency by solving dynamical equations, but it incurs prohibitive computational cost under high resolution and frequent update cycles~\cite{Bauer2021,quietrevolution}. In contrast, data-driven forecasters trained on large reanalysis datasets enable substantially faster inference while achieving competitive short-to-medium-range accuracy~\cite{graphcast,Accurate}. Despite their efficiency, purely data-driven forecasters do not explicitly enforce physical consistency across variables and space, where small short-range errors can be amplified under autoregressive deployment and evolve into physically implausible states. In contrast, operational forecasting routinely performs state-aware diagnosis and targeted corrections to maintain coherent synoptic structures.

\begin{figure}[t]
  \centering
  \includegraphics[width=\linewidth]{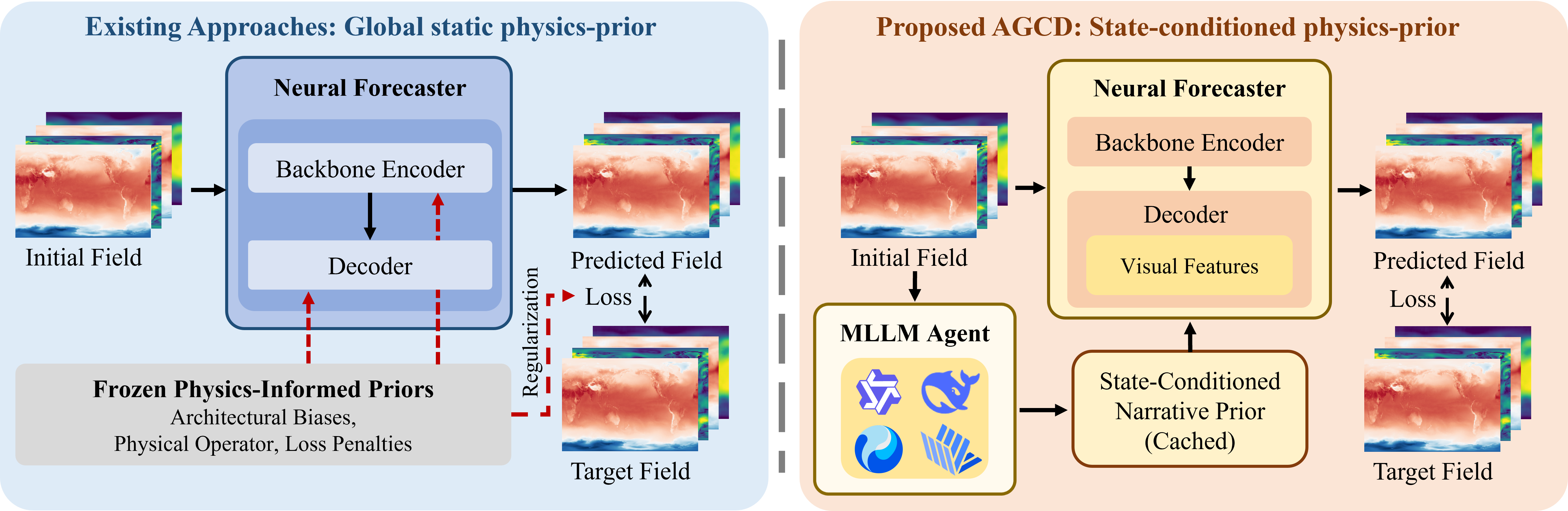}
  \captionsetup{skip=2pt} 
  \caption{Global static physics-priors vs. State-conditioned physics-priors: proposed AGCD injects cached state-conditioned physics-priors at decoding time.}
  \vspace{-1em}
  \label{fig:prior_paradigms}
\end{figure}

Recognizing that purely data-driven, grid-wise regression is insufficient to preserve meteorologically meaningful structures and constraints under complex atmospheric dynamics, prior work has revisited a central principle of NWP: constraining evolution with physical knowledge. Accordingly, researchers have attempted to inject meteorological physical priors into learning-based forecasters in various forms to guide physics-aware representation learning and improve predictive performance. 
Existing attempts to incorporate physical knowledge into data-driven forecasters mainly differ in where the prior is imposed: (1) model-level biases baked into architectures (e.g., spectral or operator designs~\cite{climODE,Neuralgeneral,fourcast}, variable embeddings~\cite{VarteX,Learningtoforecast,Triformer}, spherical/mesh representations~\cite{GlobalForecasting}, and tailored objectives~\cite{machinelearning}), (2) training-time constraints added as regularization or physics-informed objectives~\cite{Interpretable,Enforcing,NeuralNetworks,Physics-informed}, and (3) Hybrid schemes~\cite{candeep} that couple with NWP can enhance physical consistency. While effective, these priors are usually imposed in a global, once-for-all manner, limiting sample-specific controllability and state-adaptive guidance during multi-step deployment. Fig.~\ref{fig:prior_paradigms} summarizes this gap and motivates an alternative: deriving state-conditioned, physically consistent guidance from the current atmosphere and applying it in a controllable and reusable way.

Recently, Multimodal Large Language Models (MLLMs) and agent workflows have achieved strong results across computer vision~\cite{chen2024longvilascalinglongcontextvisual,wang2024qwen2vlenhancingvisionlanguagemodels,dai2026needrealanomalymllm,chen2024internvlscalingvisionfoundation,li2025vidhallucevaluatingtemporalhallucinations,tong2024cambrian1fullyopenvisioncentric,xiao2023florence2advancingunifiedrepresentation,11093955} and natural language processing~\cite{11269711,jiang2025specificmllmsomnimllmssurveymllms,chen2025multimodallanguagemodelsbetter,Yin_2024,zhang2024mmllmsrecentadvancesmultimodal,caffagni2024revolutionmultimodallargelanguage}, and are increasingly trained with an emphasis on physical consistency and guidance~\cite{PaLM-E,qwen3technicalreport,wu2024deepseekvl2mixtureofexpertsvisionlanguagemodels,hunyuanocr2025,wang2025internvl3_5}. 
Their ability to produce consistent visual descriptions suggests a route to summarize the current multi-variable atmosphere into an explicit, controllable prior that highlights synoptic structures and enforces cross-variable consistency.
Unlike static physics injection that is baked into architectures or losses and is hard to steer per sample, state-conditioned summaries provide situation-aware guidance with per-variable evidence and checkable consistency constraints.
However, naively applying generic captioners or single-round MLLMs to meteorological fields is hindered by two bottlenecks: \textbf{reliability} (coverage gaps and cross-variable inconsistencies in strongly coupled, high-dimensional states) and \textbf{efficiency} (online multi-agent reasoning is costly for training and deployment).
Therefore, we seek a reliable and efficient mechanism that generates causally valid, state-conditioned priors and injects them into forecasters with less runtime cost. 

Motivated by this perspective, we propose {Agent-Guided Cross-modal Decoding (\textbf{AGCD})}, a plug-and-play prior-injection paradigm designed for Transformer-based neural forecasters. Concretely, AGCD employs an offline {Multi-agent Meteorological Narration Pipeline (\textbf{MMNP})} to generate concise, state-conditioned physics-priors from multi-variable heatmaps, and injects them as decoding-time guidance into Transformer-based forecasters. To realize this, we further introduce {Cross-modal Region Interaction Decoding (\textbf{CRID})}, a plug-and-play cross-modal decoder that efficiently fuses the cached priors with visual tokens for region-adaptive refinement, improving structural fidelity without modifying the backbone I/O interface. We evaluate AGCD on WeatherBench~\cite{WeatherBench} at two resolutions; for long-horizon assessment, we perform strictly causal 6 hour-step autoregressive rollouts up to 48 hours, where the narrative is refreshed from the current rollout state without introducing future information. Across settings, AGCD consistently improves accuracy and reduces error accumulation, leading to more stable long-horizon behavior.

\noindent\textbf{Contributions.}
Our main contributions are three-fold:
\begin{itemize}
  \item We introduce a new perspective for physics-priors injection in weather forecasting: leveraging MLLMs to convert multi-variable atmospheric states into state-conditioned physics-priors that are explicit, controllable, and reusable.
  \item We propose AGCD, a plug-and-play decoding-time prior-injection framework that couples an offline multi-agent narration pipeline (MMNP) with a lightweight cross-modal decoder (CRID) to enable region-adaptive refinement without modifying backbone interfaces.
  \item We demonstrate consistent gains for 6-hour forecasting on WeatherBench across two resolutions and diverse backbones (generic and weather-specialized), including 48-hour autoregressive rollouts that reduce early-stage error accumulation.
\end{itemize}

\section{Related Work}

\subsection{Data-driven Weather Forecasting}
Data-driven weather forecasting has advanced rapidly with deep models that learn spatiotemporal dynamics from large reanalysis datasets.
Beyond early convolutional and recurrent approaches, recent progress mainly follows three directions:
(i) Neural operators that approximate the evolution operator in function space and enable efficient global mixing for autoregressive rollouts~\cite{fno,fourcast,sfno};
(ii) Transformer-style forecasters that scale sequence modeling on latitude--longitude grids, often incorporating weather-aware designs such as variable level embeddings, pressure-level structure, and latitude-weighted objectives~\cite{climax,pangu,Accurate};
and (iii) graph-based forecasters that perform message passing on spherical meshes for long-range transport and multi-scale interactions beyond regular grids~\cite{graphcast,ijcai2025p885,linander2025pear}.
These methods have achieved strong single-step accuracy and practical inference efficiency, making them promising alternatives or complements to traditional NWP in short-range settings.

Existing physics-priors injection is largely static, which is often hard to control at training time and lacks a mechanism for state-aware revision over dynamically sensitive regions. This static becomes fragile under autoregressive deployment: small structural misplacements and weak cross-variable coherence at early steps can be recursively amplified, yielding systematic bias and unstable long-horizon trajectories. These limitations motivate an explicit, controllable, and plug-and-play guidance mechanism that injects state-conditioned priors at decoding time, without redesigning strong backbones, thereby improving the stability of early-stage autoregressive rollouts.

\subsection{MLLMs and Agentic Workflows for Structured Guidance}
Recent multimodal large language models (MLLMs) and agentic workflows~\cite{chen2025egoagent} have become practical mechanisms for structured guidance, showing strong capabilities in grounded description~\cite{tong2024cambrian1fullyopenvisioncentric,xiao2023florence2advancingunifiedrepresentation,he2025radarqa,gao2025interleaved}, region-centric reasoning~\cite{chen2024internvlscalingvisionfoundation,wang2024qwen2vlenhancingvisionlanguagemodels}, multi-step decomposition~\cite{gupta2023visual,chen2025agentcaster,shi2025enhancing,wang2024cog,yang2025vca,ghezloo2025pathfinder}, and tool-augmented verification~\cite{caffagni2024revolutionmultimodallargelanguage,zhang2024mmllmsrecentadvancesmultimodal,varambally2026zephyrus} across vision and language tasks. In particular, verification-oriented designs are used to suppress omissions, contradictions, and overconfident statements~\cite{li2025mccd,yang2025nader,dhuliawala2024chain}. This paradigm suggests a route to convert high-dimensional visual observations~\cite{chen2025less} into compact semantic summaries that can act as controllable signals for downstream models~\cite{liao2025motionagent,chen2025t2i,yao2024promptcot}.

However, transferring generic captioners or online multi-agent reasoning to meteorological fields is challenging due to two constraints: \textbf{reliability} and \textbf{efficiency}. Weather states are high-dimensional and strongly cross-variable coupled, making one-shot generation prone to incomplete coverage and inconsistent semantics, which is undesirable as a stable training-time prior. Meanwhile, online multi-agent execution is costly and often difficult to reproduce within large-scale forecasting pipelines. These limitations motivate guidance mechanisms that produce deterministic, evidence-grounded state summaries with explicit consistency control, enable offline caching to avoid online multi-agent iterations during training and one-step inference while supporting strictly causal rollouts via a lightweight  single-step editor.
\section{Methodology}
\subsection{Overall Framework}
\label{subsec:overview}

\begin{figure*}[t]
  \begin{center}
    \centerline{\includegraphics[width=\textwidth]{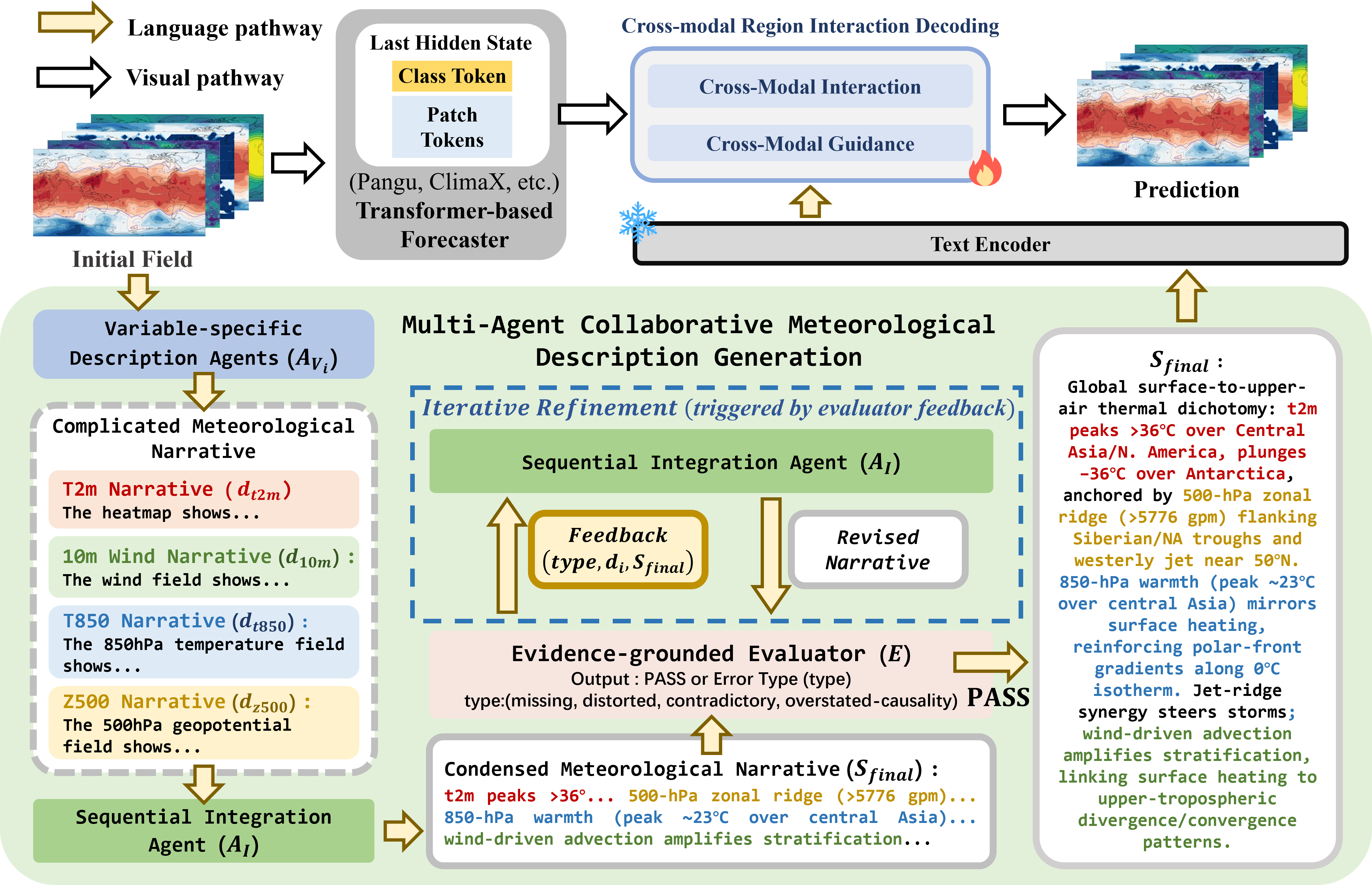}}
    \caption{
      The overview of the proposed AGCD. 
    }
    \label{fig:framework}
    \vspace{-2em}
  \end{center}
\end{figure*}
As illustrated in Fig.~\ref{fig:framework}, our framework couples structured language guidance with visual spatiotemporal representation learning for meteorological forecasting. It consists of a language pathway that provides semantic cues and a visual pathway that produces spatiotemporal tokens for prediction.

\textbf{Language pathway.} For each meteorological variable field $X_i\in\mathbb{R}^{H\times W}$, we render it into an RGB heatmap $\mathbf{I}_i\in\mathbb{R}^{H\times W\times 3}$ using a fixed colormap and a fixed normalization scheme to ensure a deterministic value-to-color mapping. Given the multivariate heatmaps $\{\mathbf{I}_i\}_{i=1}^{N}$, the proposed Multi-agent Meteorological Narration Pipeline (\textbf{MMNP}) (Sec.~\ref{subsec:MMNP}) generates a coherent meteorological narrative $S_{\text{final}}$ summarizing salient atmospheric states and potential inter-variable interactions. To avoid running multi-agent iterations, $S_{\text{final}}$ is precomputed offline for each sample and cached for training and inference.

We then encode $S_{\text{final}}$ using a pretrained Large Language Model (LLM) and extract the last-layer hidden states as token embeddings:
\begin{equation}
\mathbf{T}=\mathcal{E}_{\text{LLM}}(S_{\text{final}})\in\mathbb{R}^{N_t\times d_t}.
\end{equation}
Importantly, the LLM is kept frozen throughout training and inference.

\textbf{Visual pathway and cross-modal coupling.} In parallel, the raw fields are fed into a Transformer-based forecasting backbone (such as Pangu \cite{pangu}, ClimaX \cite{climax}, etc.), producing patch tokens $P\in\mathbb{R}^{N\times d}$ (with $N=H\cdot W$) and a global class token $C\in\mathbb{R}^{1\times d}$. We then perform a cross-modal guidance preprocessing step in our Cross-modal Region Interaction Decoding (\textbf{CRID}) (Sec.~\ref{sec:crid}). Specifically, the class token $C$ generates token-wise and channel-wise gates to refine the frozen text embeddings $T$, yielding visual-guided text features $\tilde{T}$ aligned to the visual feature space. CRID then injects $\tilde{T}$ into region-aware decoding through token distillation and cross-attention modulation, producing improved forecasts that leverage both local atmospheric patterns and global semantic context.

\subsection{Multi-agent Meteorological Narration Pipeline (MMNP)}
\label{subsec:MMNP}

Generating a coherent and meteorologically plausible narrative from multivariate atmospheric inputs requires (i) capturing salient spatial patterns within each variable and (ii) integrating cross-variable cues without introducing contradictions or temporally-confounded causal claims. Therefore, we propose \textbf{MMNP}, a collaborative multi-agent pipeline that produces an offline narrative prior $S_{\text{final}}$ from deterministically rendered RGB heatmaps $\{\mathbf{I}_i\}_{i=1}^{N}$ (Sec.~\ref{subsec:overview}). To keep computation bounded and reproducible, MMNP operates under fixed prompts templates and a fixed refinement budget.

\subsubsection{Agents and roles}
MMNP consists of three agents with complementary responsibilities:

\paragraph{(1) Variable-specific description agents $\mathbf{A}_{V_i}$.}
For each variable $V_i$, agent $A_{V_i}$ takes the corresponding heatmap $\mathbf{I}_i$ and extracts salient spatial structures with coarse localization cues in a concise textual form:
\begin{equation}
d_i = A_{V_i}(\mathbf{I}_i), \qquad i=1,\ldots,N.
\end{equation}
Each $d_i$ follows a lightweight, template-constrained style (short clauses with approximate regions and intensity trends) to facilitate downstream integration and verification.

\paragraph{(2) Sequential integration agent $\mathbf{A}_I$.}
The integration agent $A_I$ merges $\{d_i\}_{i=1}^{N}$ into a unified narrative by iteratively updating a running state $S_{i-1}\!\rightarrow S_i$ under a fixed variable order:
\begin{equation}
S_i = A_I(S_{i-1}, d_i), \qquad S_0=\emptyset.
\end{equation}
To prevent uncontrolled verbosity and to ensure consistent phrasing across samples, $A_I$ writes $S_i$ in a template-constrained format (short sentences or bullets) and explicitly separates:
(a) observations grounded in the current heatmap patterns, and
(b) hypothesized interactions across variables, phrased as tentative rather than factual or future-dependent claims.

\paragraph{(3) Evidence-grounded evaluator $E$.}
Given the variable-wise descriptions $\{d_i\}_{i=1}^{N}$ and the integrated narrative $S_{\text{final}}$, the evaluator $E$ performs a structured consistency check and returns either \texttt{PASS} or a feedback package. Concretely, $E$ assesses three aspects:
\begin{itemize}
\item \textbf{Per-variable coverage}: whether salient structures described in each $d_i$ are reflected in $S_{\text{final}}$ (mitigating coverage gaps);
\item \textbf{Consistency with described evidence}: whether statements in $S_{\text{final}}$ preserve the coarse localization and intensity trends stated in $\{d_i\}$, without distortion or unwarranted specificity;
\item \textbf{Coherence}: whether the narrative is concise, well-structured, and non-redundant.
\end{itemize}
The evaluator reports localized issues with different types (such as {missing}, {distorted}, {contradictory}, and {overstated-causality}) to enable targeted refinement.

\subsubsection{Forward generation and evaluation}
All variable-specific agents are executed in parallel to produce $\{d_i\}_{i=1}^{N}$, followed by chained integration to obtain $S_{\text{final}}$. The evaluator then verifies $S_{\text{final}}$ against the variable-wise descriptions:
\begin{equation}
\texttt{flag} = E(\{d_i\}_{i=1}^{N}, S_{\text{final}}).
\end{equation}
If \texttt{flag} is \texttt{PASS}, we output $S_{\text{final}}$ as the final narrative prior for subsequent frozen-LLM. 
If the \texttt{flag} is  \texttt{Fail}, $E$ returns a feedback package that specifies the issue type and the implicated variable, together with the current integrated narrative:

\begin{equation}
\texttt{Feedback} = (\texttt{type}, i, d_i, S_{\text{final}}),
\end{equation}

where \texttt{type} $\in$ \{missing, distorted, contradictory, overstated-causality\} and $i$ indexes the variable whose description is implicated. Conditioned on \texttt{Feedback}, the integration agent $A_I$ revises $S_{\text{final}}$ by (i) adding missing but supported content from $d_i$, (ii) correcting distorted localization and intensity phrasing, (iii) resolving contradictions by rephrasing or narrowing claims and (iv) weakening causal language into hypothesis form, while preserving unaffected content.

To ensure bounded and reproducible computation, we run at most $R$ refinement rounds (fixed across the dataset). If the narrative still fails after $R$ rounds, we fall back to the best-scoring version selected by $E$. The finalized $S_{\text{final}}$ is cached offline and reused during training and inference, avoiding online multi-agent iterations during optimization.

\subsection{Cross-modal Region Interaction Decoding (CRID)}
\label{sec:crid}
\begin{wrapfigure}[9]{r}{0.46\textwidth}
  \vspace{-10pt}
  \centering
  \includegraphics[width=\linewidth]{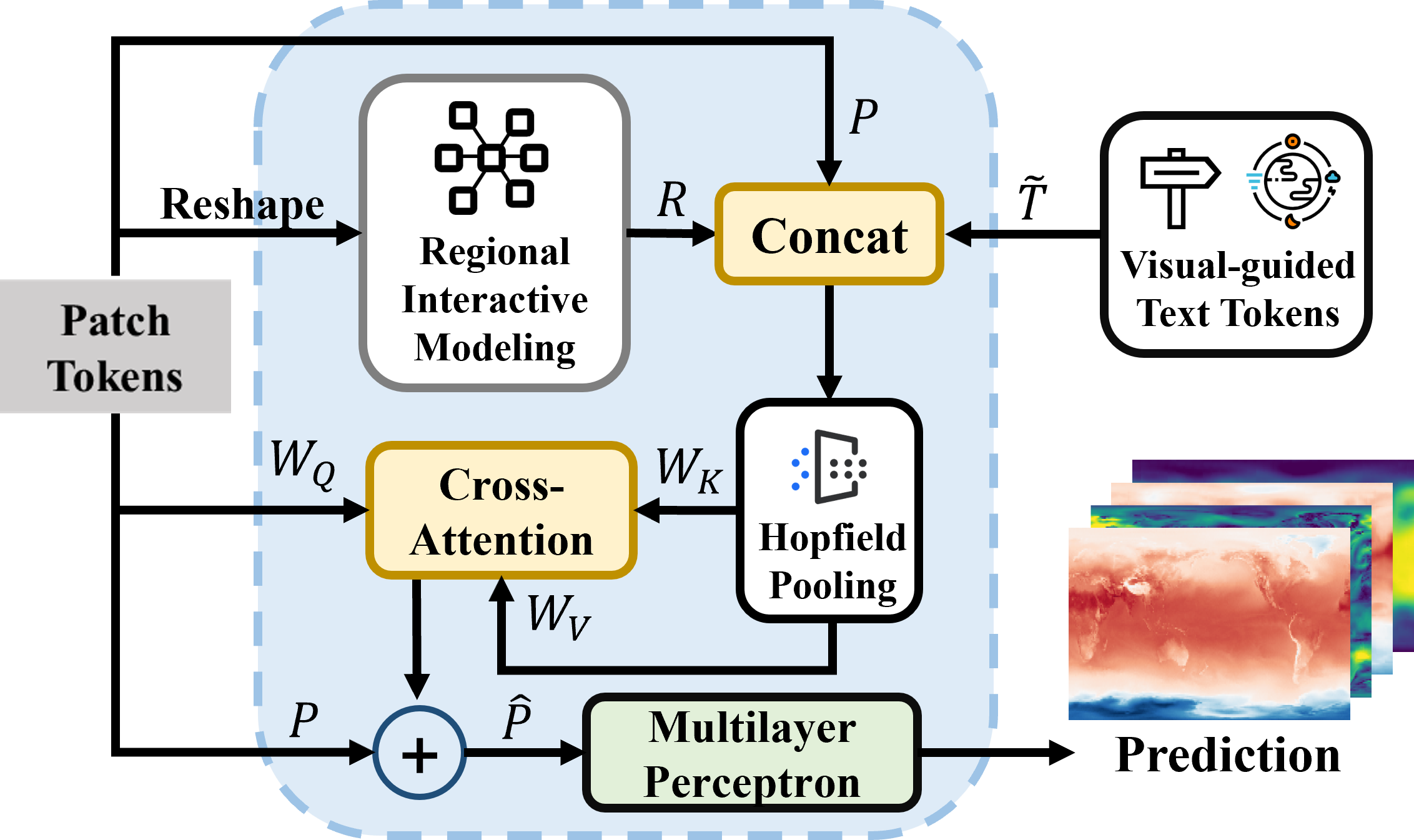}
  \captionsetup{skip=2pt}
  \caption{Structure of \textbf{Cross-Modal Interaction}.}
  \vspace{-10pt}
  \label{fig:cmi}
\end{wrapfigure}

As discussed in Sec.~\ref{subsec:overview}, the generated meteorological narrative is not merely for post-hoc explanation; instead, it should serve as a decoding-time explicit physics-priors that guides the forecaster toward dynamically sensitive regions and cross-variable-consistent structures. To this end, we propose CRID, a plug-and-play decoder that injects state-conditioned physics-priors without changing the backbone interface. CRID consists of two components: Cross-Modal Guidance (\textbf{CMG}), which produces visual-conditioned text features, and Cross-Modal Interaction (\textbf{CMI}), which performs region-aware multi-source interaction to modulate patch tokens for forecasting (See Fig.~\ref{fig:cmi}).

\paragraph{Inputs.}
Given an input state at time $t$, the forecasting backbone produces a set of patch-wise visual tokens
$P\in\mathbb{R}^{N\times d}$ (with $N=H\!\cdot\! W$) and a global summary token $C\in\mathbb{R}^{1\times d}$. In parallel, a frozen text encoder embeds the narrative prior (Sec.~\ref{subsec:overview}) into token features $T\in\mathbb{R}^{N_t\times d_t}$. CRID takes $(P,C,T)$ as inputs and performs decoding-time revision.

\subsubsection{Cross-Modal Guidance (CMG):}
\label{subsec:cmg}
CMG converts frozen text features into visual-conditioned semantics. The core idea is to use the class token $C$ as a compact summary of the current atmospheric state and let it gate the narrative tokens $T$,  thereby selectively emphasizing state-relevant semantic cues.
We first align text features to the visual channel dimension as Eq.~\eqref{eq:u}, where $g(\cdot)$ is a learnable linear projection if $d_t\neq d$. We then map $C$ through a lightweight MLP $f(\cdot)$ and split the output into two queries as Eq.~\eqref{eq:tok}:
{\setlength{\abovedisplayskip}{-5pt}
 \setlength{\belowdisplayskip}{-9pt}
 \setlength{\abovedisplayshortskip}{-9pt}
 \setlength{\belowdisplayshortskip}{4pt}

\begin{equation}
U = g(T)\in\mathbb{R}^{N_t\times d},
\label{eq:u}
\end{equation}
\begin{equation}
[q_{\text{tok}}, q_{\text{ch}}] = f(C),\quad
q_{\text{tok}}\in\mathbb{R}^{1\times N_t},\ \ q_{\text{ch}}\in\mathbb{R}^{1\times d}.
\label{eq:tok}
\end{equation}
}

Token-wise gating reweights narrative tokens by their compatibility with the global state as Eq.~\eqref{eq:alpha}, and channel-wise gating further refines the semantic channels to match the state-dependent emphasis as Eq.~\eqref{eq:beta}:

{\setlength{\abovedisplayskip}{-5pt}
 \setlength{\belowdisplayskip}{-9pt}
 \setlength{\abovedisplayshortskip}{-9pt}
 \setlength{\belowdisplayshortskip}{4pt}

\begin{equation}
\alpha = \mathrm{softmax}(q_{\text{ch}}U^\top)\in\mathbb{R}^{1\times N_t},\quad
U^{(1)} = \alpha\odot U.
\label{eq:alpha}
\end{equation}
\begin{equation}
\beta = \mathrm{softmax}(q_{\text{tok}}U^{(1)})\in\mathbb{R}^{1\times d},\quad
\tilde{T}=\beta\odot U^{(1)}.
\label{eq:beta}
\end{equation}
}

The resulting $\tilde{T}\in\mathbb{R}^{N_t\times d}$ serves as a {state-conditioned physics-priors} and will be injected into CMI for region-aware interaction.

\subsubsection{Cross-Modal Interaction (CMI):}
\label{subsec:cmi}

CMI injects the guided semantics $\tilde{T}$ into patch tokens via region-aware tokenization and memory-based modulation. Given patch tokens $P\in\mathbb{R}^{N\times d}$, we construct multi-scale region tokens by pooling on the token grid. Let $P_{\text{grid}}\in\mathbb{R}^{H\times W\times d}$ be the reshaped tokens, and for scales $\mathcal{S}$ we compute
\begin{equation}
R^{(s)}=\mathrm{Flatten}\!\left(\mathrm{AvgPool}_{s\times s}(P_{\text{grid}})\right),\quad
R=[R^{(s)}]_{s\in\mathcal{S}}\in\mathbb{R}^{N_r\times d}.
\end{equation}
We first construct a unified decoding context by concatenating patch tokens, multi-scale region tokens, and guided semantic tokens:
\begin{equation}
X=\mathrm{Concat}(P, R, \tilde{T})=[P;\,R;\,\tilde{T}]\in\mathbb{R}^{L\times d},\quad L=N+N_r+N_t.
\end{equation}
Since directly operating on $X$ is computationally expensive and may dilute salient cross-modal cues, we further distill $X$ into a compact set of $M$ memory tokens ($M\ll L$) via Hopfield pooling~\cite{ramsauer2020hopfield}, yielding representative prototypes for efficient decoding-time modulation:
\begin{equation}
Z=\mathrm{HopfieldPool}(Q_h, X)\in\mathbb{R}^{M\times d},
\end{equation}
where $Q_h\in\mathbb{R}^{M\times d}$ denotes learnable pooling queries. 
We apply multi-head attention (MHA) with $P$ as queries and the memory $Z$ as keys and values:
\begin{equation}
\hat{P}=\mathrm{MHA}(PW_Q, ZW_K, ZW_V)+P,\qquad
P_{\text{out}}=\mathrm{MLP}(\hat{P}).
\end{equation}
where $W_Q, W_K, W_V$ are learnable linear projections for queries, keys, and values, respectively.
The proposed CMI acts as a plug-in decoder that replaces the original decoding head and directly outputs the final forecasts, without modifying the backbone encoder.

\section{Experiments}

\subsection{Setup}
\paragraph{\textbf{Dataset.}}
 We evaluate on WeatherBench at 5.625$^\circ$ and 1.40625$^\circ$ for 6 hour forecasting: given state at time $t$, predict $t{+}6\mathrm{h}$. We further assess long-horizon behavior via autoregressive rollouts up to 48 hours by iteratively feeding predictions back as inputs. Inputs include surface variables $\{\texttt{wind10m}, \texttt{t2m}\}$ and upper-air variables $\{\texttt{z}, \texttt{r}, \texttt{q}, \texttt{wind}, \texttt{t}\}$ over 13 pressure levels; we report canonical WeatherBench scores on Z500, T850, T2m, and 10m wind. We use a strict temporal split: train (1979-01-01 to 2016-12-31) and test (2017-01-01 to 2017-12-31).

\paragraph{\textbf{Metrics.}}
All methods are trained under an identical supervised setup to predict $t{+}6\mathrm{h}$ from $t$, using the same variable configuration and optimization schedule; evaluation uses latitude-weighted RMSE and ACC computed on climatology-based anomalies. Full details are provided in the supplementary material (Sec.~S2).

\paragraph{\textbf{Baselines.}} 
We evaluate AGCD as a plug and play module on bothgeneric vision backbones and weather-specialized forecasters. 
\textbf{ViT}~\cite{VIT} is a pure Transformer that models an image as a sequence of patch tokens, serving as a strong and scalable generic backbone for grid-like inputs. 
\textbf{CaiT}~\cite{Cait} extends ViT by introducing class-attention mechanisms to enable deeper Image Transformers with improved optimization and representation.
\textbf{ClimaX}~\cite{climax} is a foundation model for weather and climate that is designed to be flexible over heterogeneous datasets (different variables and spatiotemporal coverage), and can be pretrained and then finetuned for downstream forecasting tasks.
\textbf{Pangu-Weather}~\cite{pangu} is a high-resolution global weather forecasting model that performs fast deterministic forecasts with a 3D architecture tailored to atmospheric fields.

\paragraph{\textbf{Implementation Details}}
MMNP uses fixed prompt templates with a bounded refinement budget $R$ to produce deterministic physics-priors from multi-variable heatmaps. Full MMNP details and all hyperparameters are provided in the supplementary material (Sec.~S1--S2; Table~S1).

\subsection{6 hour Forecasting}
\begin{figure}[t]
    \centering
    \includegraphics[width=\columnwidth]{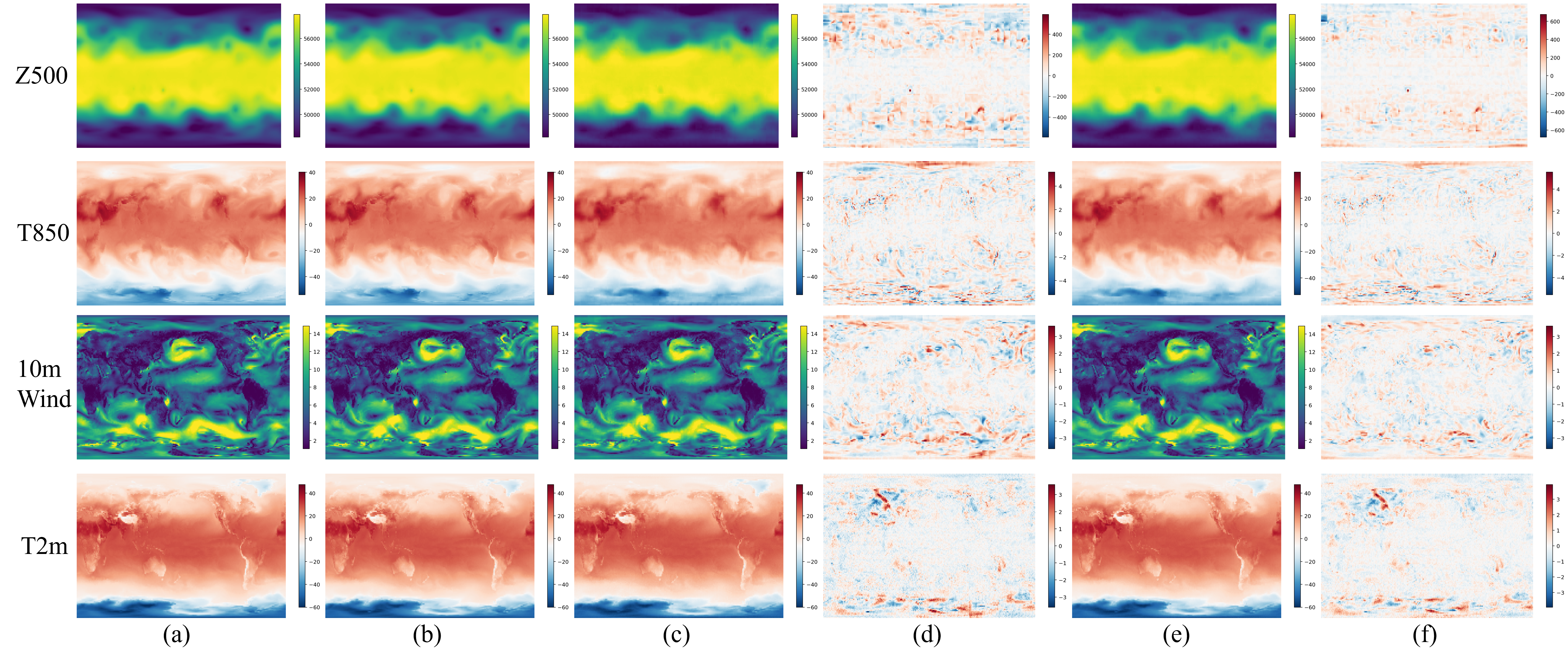}
    \caption{
        Qualitative comparison of 6 hour weather forecasting with Pangu and Pangu+AGCD (AGCD) on 1.40625$^\circ$ data across multiple variables.
        (a) Initial fields at time $t$.
        (b) Ground-truth targets at $t{+}6$h.
        (c) Predictions from the vanilla Pangu.
        (d) Error maps from the vanilla Pangu.
        (e) Predictions from Pangu with our AGCD.
        (f) Error maps from Pangu with our AGCD.
        Error maps visualize Pred$-$GT.
    }
    \label{fig:pangu_agent_qual}
    \vspace{-1em}
\end{figure}

For each framework, we report both the vanilla model and with \textbf{AGCD} counterpart obtained by plugging our semantic guidance (MMNP+CRID) into the decoding stage.
Tables~\ref{tab:main_results_two_res} summarize the 6 hour forecasting performance at 5.625$^\circ$ and 1.40625$^\circ$, respectively. Our proposed plug-and-play AGCD consistently improves all tested backbones, reducing RMSE and increasing ACC on the canonical variables.
We provide qualitative comparisons of representative 6 hour forecasts for Z500, T850, T2m, and 10m wind at 1.40625$^\circ$ (Pangu Figs.~\ref{fig:pangu_agent_qual}) and 5.625$^\circ$ (ClimaX), respectively, which shows that our proposed method yields results that closely match the ground with smaller bias. The 5.625$^\circ$ visualization is deferred to the supplementary material (Sec.~S3).

\begin{table}[t]
\centering
\small
\caption{6-hour forecasting results on WeatherBench at two resolutions. AGCD consistently improves RMSE and ACC across backbones. RMSE $\downarrow$ and ACC $\uparrow$.}
\vspace{-1em}
\label{tab:main_results_two_res}
\resizebox{\textwidth}{!}{%
\begin{tabular}{l|cccc|cccc|cccc|cccc}
\toprule
\multirow{3}{*}{Method}
& \multicolumn{4}{c|}{T2m [K]}
& \multicolumn{4}{c|}{10m Wind [m/s]}
& \multicolumn{4}{c|}{Z500 [m$^2$/s$^2$]}
& \multicolumn{4}{c}{T850 [K]} \\
\cmidrule(lr){2-5}\cmidrule(lr){6-9}\cmidrule(lr){10-13}\cmidrule(lr){14-17}
& \multicolumn{2}{c}{5.625$^\circ$} & \multicolumn{2}{c|}{1.40625$^\circ$}
& \multicolumn{2}{c}{5.625$^\circ$} & \multicolumn{2}{c|}{1.40625$^\circ$}
& \multicolumn{2}{c}{5.625$^\circ$} & \multicolumn{2}{c|}{1.40625$^\circ$}
& \multicolumn{2}{c}{5.625$^\circ$} & \multicolumn{2}{c}{1.40625$^\circ$} \\
\cmidrule(lr){2-3}\cmidrule(lr){4-5}
\cmidrule(lr){6-7}\cmidrule(lr){8-9}
\cmidrule(lr){10-11}\cmidrule(lr){12-13}
\cmidrule(lr){14-15}\cmidrule(lr){16-17}
& RMSE $\downarrow$ & ACC $\uparrow$ & RMSE $\downarrow$ & ACC $\uparrow$
& RMSE $\downarrow$ & ACC $\uparrow$ & RMSE $\downarrow$ & ACC $\uparrow$
& RMSE $\downarrow$ & ACC $\uparrow$ & RMSE $\downarrow$ & ACC $\uparrow$
& RMSE $\downarrow$ & ACC $\uparrow$ & RMSE $\downarrow$ & ACC $\uparrow$ \\
\midrule
ViT
& 1.6859 & 0.9554 & 1.3570 & 0.9710
& 0.5490 & 0.9723 & 0.5946 & 0.9668
& 131.01 & 0.9929 & 80.80 & 0.9970
& 1.21 & 0.9736 & 0.91 & 0.9852 \\
ViT+AGCD
& 1.2601 & 0.9768 & 1.2450 & 0.9754
& 0.4781 & 0.9788 & 0.5600 & 0.9695
& 113.07 & 0.9951 & 75.90 & 0.9976
& 0.98 & 0.9830 & 0.86 & 0.9871 \\
\midrule
CaiT
& 1.8747 & 0.9317 & 1.5200 & 0.9658
& 0.6051 & 0.9674 & 0.6420 & 0.9622
& 149.95 & 0.9917 & 104.60 & 0.9953
& 1.51 & 0.9516 & 1.06 & 0.9829 \\
CaiT+AGCD
& 1.8703 & 0.9450 & 1.4700 & 0.9682
& 0.5993 & 0.9678 & 0.6210 & 0.9638
& 132.80 & 0.9938 & 96.80 & 0.9960
& 1.41 & 0.9635 & 0.99 & 0.9846 \\
\midrule
ClimaX
& 1.2308 & 0.9759 & 0.7799 & 0.9904
& 0.4970 & 0.9776 & 0.3443 & 0.9889
& 88.98  & 0.9972 & 32.84 & 0.9995
& 0.92 & 0.9857 & 0.49 & 0.9957 \\
ClimaX+AGCD
& 0.8843 & 0.9880 & 0.7420 & 0.9912
& 0.4513 & 0.9812 & 0.3320 & 0.9896
& 70.22  & 0.9979 & 31.10 & 0.9996
& 0.78 & 0.9905 & 0.46 & 0.9962 \\
\midrule
Pangu
& 0.4965 & 0.9961 & 0.5147 & 0.9958
& 0.5963 & 0.9666 & 0.5321 & 0.9733
& 80.91  & 0.9970 & 74.36 & 0.9974
& 0.52 & 0.9951 & 0.67 & 0.9920 \\
Pangu+AGCD
& 0.4916 & 0.9962 & 0.4551 & 0.9967
& 0.5507 & 0.9716 & 0.4451 & 0.9814
& 68.92  & 0.9978 & 58.73 & 0.9984
& 0.50 & 0.9954 & 0.63 & 0.9929 \\
\bottomrule
\end{tabular}%
}
\end{table}

\subsection{Autoregressive forecasting}
\paragraph{\textbf{Text update rule for rollouts.}}
While our base task is 6 hour forecasting and the narrative prior is intentionally concise, regenerating the full MMNP at every rollout step is unnecessary and inefficient. Therefore, we adopt a lightweight rollout update: we keep the variable-specific describers and evaluator off during rollouts, and reuse only the sequential integration agent as a single-step editor. In all autoregressive experiments, we instantiate this editor with InternVL3.5. Concretely, at step $k$ the editor takes (i) the current predicted meteorological heatmap stack $\{\mathbf{I}_i^{(k)}\}$ and (ii) the previous-step narrative $S^{(k-1)}$, then outputs an updated narrative $S^{(k)}$ by making minimal, evidence-grounded edits:
\begin{equation}
S^{(k)} = A_I\!\left(S^{(k-1)}, \{\mathbf{I}_i^{(k)}\}_{i=1}^{N}\right).
\end{equation}
This yields a causally valid, step-adaptive physics-priors with negligible overhead, while avoiding repeated multi-agent refinement. The updated $S^{(k)}$ is then encoded by the frozen LLM and injected by CRID for the next rollout step.

\begin{figure}[t]
  \centering
  \includegraphics[width=\linewidth]{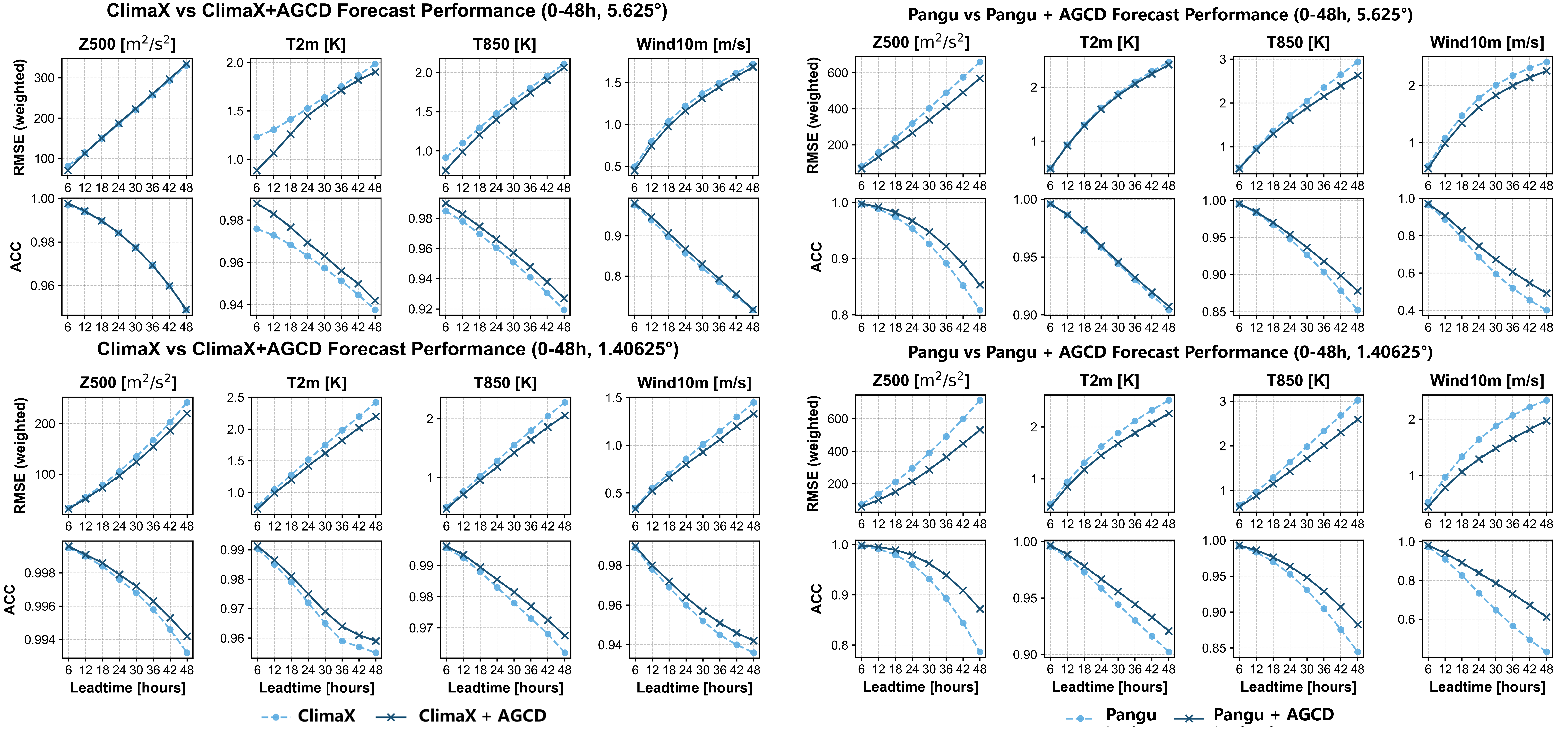}
  \captionsetup{skip=2pt} 
  \caption{Autoregressive rollout comparison between Pangu and Pangu+AGCD up to 48 hours (6 hour steps).}
  \vspace{-1em}
  \label{fig:supp_pangu_agent_ar}
\end{figure}

We evaluate AGCD via strictly causal autoregressive rollouts with a 6 hour step: starting from the initial state at time $t$, the model iteratively feeds its own prediction back as input to forecast $t{+}6h,\ldots,t{+}48h$. Fig.~\ref{fig:supp_pangu_agent_ar} reports the lead-time curves of latitude-weighted RMSE/ACC, and the detailed RMSE results at 12-hour intervals across backbones are deferred to the supplementary material (Sec.~S3). Across variables, our AGCD consistently reduces error accumulation and yields more stable trajectories under rollout.

\section{Discussion}
\label{sec:analysis}



\subsection{How crucial is semantic alignment for  improvement?}
\begin{figure}[t]
  \centering
  \includegraphics[width=\textwidth]{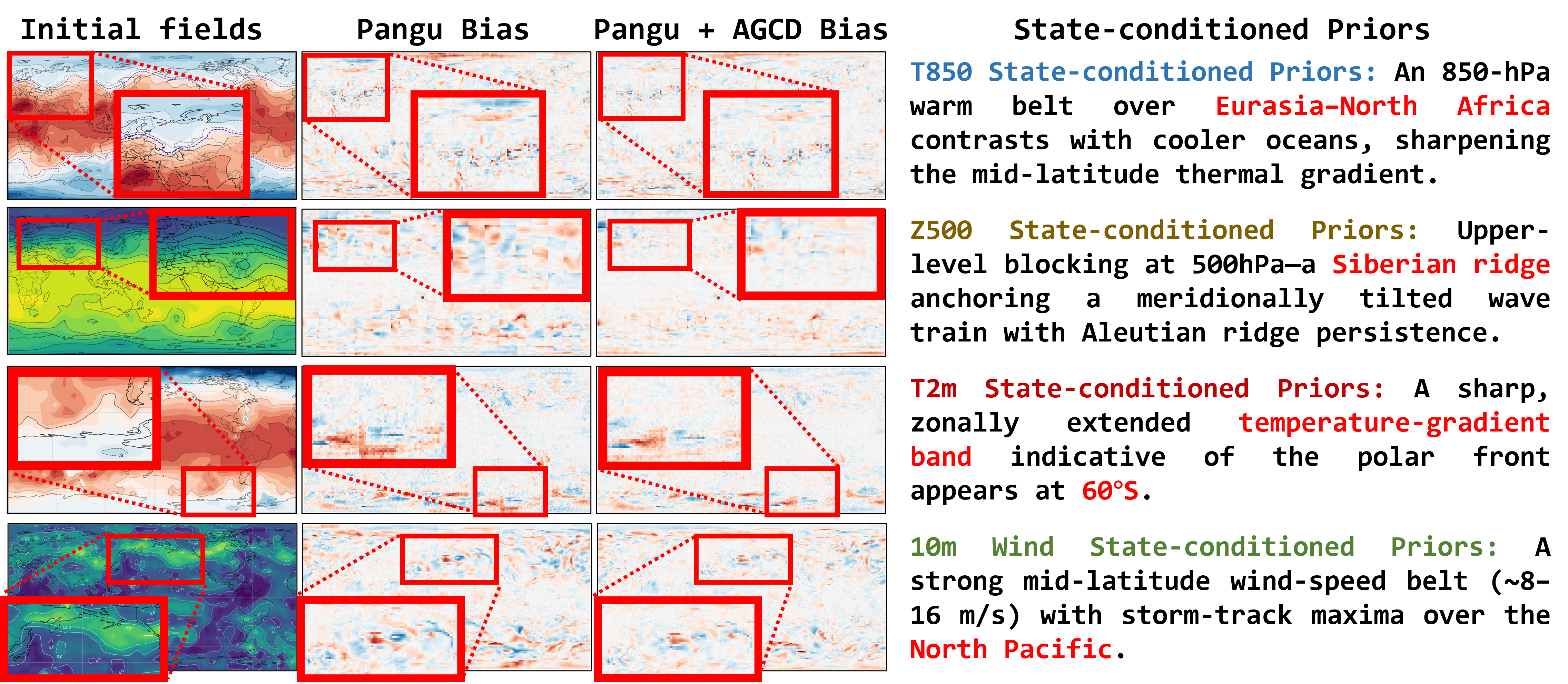}
  \caption{Relevance case study: State-consistent priors yield targeted error reductions.}
  \vspace{-1em}
  \label{fig:relevance_case}
\end{figure}

We keep the backbone and CRID identical and only change the text: \textbf{Matched} (sample-aligned), \textbf{Shuffled} (mismatched), and \textbf{Empty} (null).
Table~\ref{tab:semantic_controls} shows that improvements appear only with Matched text, while Shuffled/Empty largely remove the benefit and can even underperform the vision-only baseline, confirming that semantic alignment is necessary. Fig.~\ref{fig:relevance_case} provides a concrete example showing that matched narratives offer localized, state-consistent priors rather than generic text cues. 
For T850, the narrative explicitly highlights the dynamically active regions over Eurasia--North and Africa, which coincide with the boxed areas where the baseline exhibits structured warm/cold displacement errors. 
For Z500, the prior emphasizes the Siberian ridge, aligning with the synoptic-scale height pattern and guiding corrections on the corresponding ridge-related error patches. 
For T2m, the narrative points to a temperature-gradient band around $60^\circ$S, matching the sharp frontal-like transitions where the baseline tends to blur gradients and incur coherent bias. 
For 10m wind, the prior focuses on the North Pacific, consistent with the prominent wind structures and the concentrated error clusters in that region. 
Across variables, these region-specific priors translate into targeted error reductions in the zoomed-in boxes, supporting that the gain comes from sample-aligned semantic guidance rather than extra text capacity.
\begin{table}[t]
\centering
\caption{Semantic relevance controls. All settings keep the visual backbone (ViT) and CRID identical; only the text input is modified.}
\label{tab:semantic_controls}
\scriptsize
\setlength{\tabcolsep}{2.6pt}
\renewcommand{\arraystretch}{1.05}
\resizebox{0.92\columnwidth}{!}{%
\begin{tabular}{l|cc|cc|cc|cc}
\toprule
\multirow{2}{*}{Text setting} 
& \multicolumn{2}{c|}{Z500} 
& \multicolumn{2}{c|}{T850} 
& \multicolumn{2}{c|}{T2m} 
& \multicolumn{2}{c}{10m Wind} \\
& RMSE $\downarrow$ & ACC $\uparrow$ 
& RMSE $\downarrow$ & ACC $\uparrow$ 
& RMSE $\downarrow$ & ACC $\uparrow$ 
& RMSE $\downarrow$ & ACC $\uparrow$ \\
\midrule
Vision-only (no text) & 131.01 & 0.9929 & 1.21 & 0.9736 & 1.6859 & 0.9554 & 0.5490 & 0.9723\\
\midrule
Matched (ours)        & 113.07 & 0.9951 & 0.98 & 0.9830 & 1.2601 & 0.9768 & 0.4781 & 0.9788\\
Shuffled (mismatch)   & 136.40 & 0.9922 & 1.24 & 0.9730 & 1.7120 & 0.9550 & 0.5650 & 0.9718 \\
Empty (null prompt)   & 134.80 & 0.9924 & 1.23 & 0.9732 & 1.7050 & 0.9552 & 0.5600 & 0.9720 \\
\bottomrule
\end{tabular}%
}
\end{table}

\subsection{Can multi-agent decomposition enhance narrative reliability?}
We compare three narrative generation strategies under the same forecasting backbone and CRID:
(1) Single-agent uses only the integration agent $A_I$ to produce a single-pass narrative by taking the full multi-variable heatmap stack $\{\mathbf{I}_i\}_{i=1}^{N}$ as input, without variable-wise decomposition or post-hoc verification.
(2) Multi-agent w/o evaluator decomposes the input into variable-specific descriptions $\{d_i\}$ and integrates them with $A_I$, but removes the evidence-grounded evaluator $E$.
(3) Full MMNP further adds $E$ to detect and revise omissions and cross-variable inconsistencies.
Representative narrative comparisons are provided in the supplementary material (Sec.~S3).

Table~\ref{tab:MMNP_ablation} shows a monotonic improvement as the generation pipeline becomes more reliable:
variable-wise decomposition already improves over Single-agent, and adding the evaluator yields further gains.
This supports that the benefit comes from better evidence coverage and consistency control, rather than increasing text length.

\begin{table}[t]
\captionsetup{skip=2pt}
\centering
\small
\setlength{\tabcolsep}{3pt}
\renewcommand{\arraystretch}{1.05}
\caption{Ablation on MMNP generation strategies (same CRID and backbone (ViT)).}
\label{tab:MMNP_ablation}
\resizebox{0.92\columnwidth}{!}{%
\begin{tabular}{l|cc|cc|cc|cc}
\toprule
\multirow{2}{*}{Text generator} 
& \multicolumn{2}{c|}{Z500} 
& \multicolumn{2}{c|}{T850} 
& \multicolumn{2}{c|}{T2m} 
& \multicolumn{2}{c}{10m Wind} \\
& RMSE $\downarrow$ & ACC $\uparrow$ 
& RMSE $\downarrow$ & ACC $\uparrow$ 
& RMSE $\downarrow$ & ACC $\uparrow$ 
& RMSE $\downarrow$ & ACC $\uparrow$ \\
\midrule
Single-agent ($A_I$ only)          & 123.80 & 0.9937 & 1.05 & 0.9802 & 1.3900 & 0.9688 & 0.5070 & 0.9756 \\
Multi-agent w/o evaluator ($\{A_{V_i}\}{+}A_I$) & 118.20 & 0.9944 & 1.01 & 0.9819 & 1.3200 & 0.9742 & 0.4920 & 0.9773 \\
Full MMNP (ours) ($\{A_{V_i}\}{+}A_I{+}E$)      & 113.07 & 0.9951 & 0.98 & 0.9830 & 1.2601 & 0.9768 & 0.4781 & 0.9788 \\
\bottomrule
\end{tabular}%
}
\vspace{-1em}
\end{table}

\subsection{What drives per-variable fidelity?}
\begin{figure}[t]
  \centering
  \includegraphics[width=\linewidth]{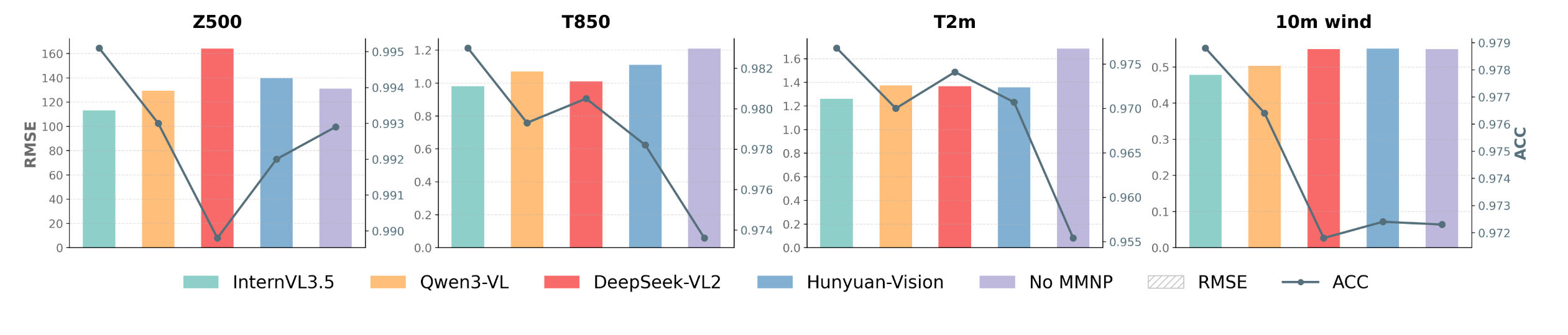}
  \captionsetup{skip=2pt}
  \caption{Ablation on variable-specific description agents in MMNP (ViT backbone; $A_I$ and $E$ fixed). We report latitude-weighted RMSE (bars; lower is better) and ACC (line; higher is better) on Z500, T850, T2m, and 10m wind.}
  \label{fig:var_agent_compare}
  \vspace{-1em}
\end{figure}

Finally, fixing the integration agent $A_I$ and evaluator $E$, we study per-variable fidelity from two angles:
(i) swapping the variable-specific agents $\{A_{V_i}\}$ while keeping the rest of the pipeline unchanged (Fig.~\ref{fig:var_agent_compare});
and (ii) incrementally enabling a subset of $\{A_{V_i}\}$ (Table~\ref{tab:MMNP_agent_enable_vit}).
Fig.~\ref{fig:var_agent_compare} shows that stronger variable-specific describers yield consistently better RMSE/ACC across all four canonical variables, confirming that MMNP benefits from higher-quality per-variable evidence rather than simply increasing text capacity.

\begin{table}[t]
\centering
\caption{Incremental ablation on enabling variable-specific description agents in MMNP (ViT backbone; $A_I$ and $E$ fixed). RMSE $\downarrow$ / ACC $\uparrow$ for 6 hour forecasts.}
\label{tab:MMNP_agent_enable_vit}
\scriptsize
\setlength{\tabcolsep}{2pt}
\renewcommand{\arraystretch}{1.08}
\begin{tabular}{p{2.35cm}|cc|cc|cc|cc}
\toprule
\multirow{2}{*}{Enabled $A_{\cdot}$} &
\multicolumn{2}{c|}{Z500} &
\multicolumn{2}{c|}{T850} &
\multicolumn{2}{c|}{T2m} &
\multicolumn{2}{c}{10m wind} \\
& RMSE $\downarrow$ & ACC $\uparrow$ &
  RMSE $\downarrow$ & ACC $\uparrow$ &
  RMSE $\downarrow$ & ACC $\uparrow$ &
  RMSE $\downarrow$ & ACC $\uparrow$ \\
\midrule
No MMNP                         & 131.01 & 0.9929 & 1.21 & 0.9736 & 1.6859 & 0.9554 & 0.5490 & 0.9723\\
$A_{\text{t2m}}$                & 130.62 & 0.9929 & 1.16 & 0.9752 & 1.5438 & 0.9620 & 0.5352 & 0.9735 \\
$A_{\text{t2m,10m}}$            & 130.12 & 0.9930 & 1.13 & 0.9768 & 1.4820 & 0.9650 & 0.5249 & 0.9745 \\
$A_{\text{t2m,10m,t850}}$       & 129.66 & 0.9930 & 1.10 & 0.9782 & 1.4187 & 0.9678 & 0.5146 & 0.9755 \\
$A_{\text{t2m,10m,t850,z500}}$  & 113.07 & 0.9951 & 0.98 & 0.9830 & 1.2601 & 0.9768 & 0.4781 & 0.9788 \\
\bottomrule
\end{tabular}
\vspace{-5pt}
\end{table}

\subsection{Which CRID components matter most?}
\label{sec:crid_ablation}

We next validate the design choices in CRID by ablating its key components while keeping the backbone, training protocol, and text generator fixed. We consider removing (i) the \textbf{region-aware multi-scale tokens}, (ii) the \textbf{Hopfield-based distillation} and directly performing attention over the concatenated tokens, and (iii) the \textbf{CMG gating} that produces visually aligned text features. Results in Table~\ref{tab:crid_ablation} indicate that each component contributes to the final performance. Notably, region-aware tokens consistently improve variables characterized by sharp gradients or coherent synoptic structures, while Hopfield distillation provides a favorable accuracy--efficiency trade-off by retaining informative cross-modal cues with reduced token complexity. Removing CMG gating degrades performance, suggesting that coarse global visual context (class token) is beneficial for reweighting and aligning frozen text embeddings before fusion.

\begin{table}[!t]
\setlength{\belowcaptionskip}{1pt} 
\centering
\caption{Ablation on CRID components.}
\label{tab:crid_ablation}
\scriptsize
\setlength{\tabcolsep}{2.6pt}
\renewcommand{\arraystretch}{1.05}
\resizebox{0.92\columnwidth}{!}{%
\begin{tabular}{l|cc|cc|cc|cc}
\toprule
\multirow{2}{*}{Setting} 
& \multicolumn{2}{c|}{Z500} 
& \multicolumn{2}{c|}{T850} 
& \multicolumn{2}{c|}{T2m} 
& \multicolumn{2}{c}{10m Wind} \\
& RMSE $\downarrow$ & ACC $\uparrow$ 
& RMSE $\downarrow$ & ACC $\uparrow$ 
& RMSE $\downarrow$ & ACC $\uparrow$ 
& RMSE $\downarrow$ & ACC $\uparrow$ \\
\midrule
Vision-only (no CRID)     & 131.01 & 0.9929 & 1.21 & 0.9736 & 1.6859 & 0.9554 & 0.5490 & 0.9723\\
+ Text, w/o Region tokens & 124.70 & 0.9936 & 1.09 & 0.9786 & 1.4200 & 0.9681 & 0.5150 & 0.9750 \\
+ Text, w/o HopfieldPool  & 115.60 & 0.9948 & 1.00 & 0.9824 & 1.3000 & 0.9758 & 0.4860 & 0.9780 \\
+ Text, w/o CMG gating    & 118.90 & 0.9943 & 1.02 & 0.9818 & 1.3400 & 0.9748 & 0.4930 & 0.9775\\
Full CRID (ours)          & 113.07 & 0.9951 & 0.98 & 0.9830 & 1.2601 & 0.9768 & 0.4781 & 0.9788 \\
\bottomrule
\end{tabular}%
}
\vspace{-10pt}
\end{table}

\section{Conclusions}
In this work, we propose \textbf{AGCD}, an explicit and plug-and-play decoding-time prior injection paradigm for neural weather forecasting. 
AGCD is motivated by a key gap in existing forecasters: grid-wise regression alone lacks a state-aware physics-priors, so structural errors and cross-variable inconsistencies can be amplified under autoregressive rollouts. 
To bridge this gap, we introduce a \textbf{MMNP} to produce state-conditioned physics-priors with evidence-grounded consistency control, and a lightweight \textbf{CRID} decoder to inject these priors for region-adaptive refinement without changing the backbone interface. 
Extensive experiments on WeatherBench at \mbox{$5.625^\circ$} and \mbox{$1.40625^\circ$} demonstrate consistent gains in latitude-weighted RMSE and ACC across both generic vision backbones and weather-specialized forecasters, and improved stability under strictly causal 48-hour autoregressive rollouts. 
Further analyses and ablations validate that the improvements stem from matched and reliable narratives with sufficient per-variable coverage, rather than merely adding extra text tokens. 
In the future, we plan to extend AGCD to broader variable sets and higher-resolution forecasting, explore more efficient state update and caching strategies for long-horizon deployment, and integrate stronger physically grounded constraints to further enhance robustness in operational settings.

\bibliographystyle{splncs04}
\bibliography{main}

@String(NeurIPS = {Adv. Neural Inform. Process. Syst.})

@String(NeurIPS = {NeurIPS})

@article{fno,
  title={Fourier neural operator for parametric partial differential equations},
  author={Li, Zongyi and Kovachki, Nikola and Azizzadenesheli, Kamyar and Liu, Burigede and Bhattacharya, Kaushik and Stuart, Andrew and Anandkumar, Anima},
  journal={arXiv preprint arXiv:2010.08895},
  year={2020}
}

@article{fourcast,
  title={Fourcastnet: A global data-driven high-resolution weather model using adaptive fourier neural operators},
  author={Pathak, Jaideep and Subramanian, Shashank and Harrington, Peter and Raja, Sanjeev and Chattopadhyay, Ashesh and Mardani, Morteza and Kurth, Thorsten and Hall, David and Li, Zongyi and Azizzadenesheli, Kamyar and others},
  journal={arXiv preprint arXiv:2202.11214},
  year={2022}
}

@inproceedings{sfno,
  title={Spherical fourier neural operators: Learning stable dynamics on the sphere},
  author={Bonev, Boris and Kurth, Thorsten and Hundt, Christian and Pathak, Jaideep and Baust, Maximilian and Kashinath, Karthik and Anandkumar, Anima},
  booktitle={International conference on machine learning},
  pages={2806--2823},
  year={2023},
  organization={PMLR}
}

@article{climax,
  title={Climax: A foundation model for weather and climate},
  author={Nguyen, Tung and Brandstetter, Johannes and Kapoor, Ashish and Gupta, Jayesh K and Grover, Aditya},
  journal={arXiv preprint arXiv:2301.10343},
  year={2023}
}

@article{pangu,
  title={Pangu-weather: A 3d high-resolution model for fast and accurate global weather forecast},
  author={Bi, Kaifeng and Xie, Lingxi and Zhang, Hengheng and Chen, Xin and Gu, Xiaotao and Tian, Qi},
  journal={arXiv preprint arXiv:2211.02556},
  year={2022}
}

@article{graphcast,
  title={Learning skillful medium-range global weather forecasting},
  author={Lam, Remi and Sanchez-Gonzalez, Alvaro and Willson, Matthew and Wirnsberger, Peter and Fortunato, Meire and Alet, Ferran and Ravuri, Suman and Ewalds, Timo and Eaton-Rosen, Zach and Hu, Weihua and others},
  journal={Science},
  volume={382},
  number={6677},
  pages={1416--1421},
  year={2023},
  publisher={American Association for the Advancement of Science}
}

@article{WeatherBench,
  title={WeatherBench: a benchmark data set for data-driven weather forecasting},
  author={Rasp, Stephan and Dueben, Peter D and Scher, Sebastian and Weyn, Jonathan A and Mouatadid, Soukayna and Thuerey, Nils},
  journal={Journal of Advances in Modeling Earth Systems},
  volume={12},
  number={11},
  pages={e2020MS002203},
  year={2020},
  publisher={Wiley Online Library}
}

@article{Accurate,
  title={Accurate medium-range global weather forecasting with 3D neural networks},
  author={Bi, Kaifeng and Xie, Lingxi and Zhang, Hengheng and Chen, Xin and Gu, Xiaotao and Tian, Qi},
  journal={Nature},
  volume={619},
  number={7970},
  pages={533--538},
  year={2023},
  publisher={Nature Publishing Group UK London}
}

@article{11269711,
  title={Incorporating LLM Versus LLM Into Multimodal Chain-of-Thought for Fine-Grained Evidence Generation},
  author={Xing, Shuyuan and He, Yao and Chen, Hanchao and Ke, Wenjun},
  journal={IEEE Access},
  volume={13},
  pages={202143--202170},
  year={2025},
  publisher={IEEE}
}

@inproceedings{jiang2025specificmllmsomnimllmssurveymllms,
  title={From specific-MLLMs to OMNI-MLLMs: a survey on MLLMs aligned with multi-modalities},
  author={Jiang, Shixin and Liang, Jiafeng and Wang, Jiyuan and Dong, Xuan and Chang, Heng and Yu, Weijiang and Du, Jinhua and Liu, Ming and Qin, Bing},
  booktitle={Findings of the Association for Computational Linguistics: ACL 2025},
  pages={8617--8652},
  year={2025}
}

@inproceedings{chen2025multimodallanguagemodelsbetter,
  title={Multimodal Language Models See Better When They Look Shallower},
  author={Chen, Haoran and Lin, Junyan and Chen, Xinghao and Fan, Yue and Dong, Jianfeng and Jin, Xin and Su, Hui and Fu, Jinlan and Shen, Xiaoyu},
  booktitle={Proceedings of the 2025 Conference on Empirical Methods in Natural Language Processing},
  pages={6688--6706},
  year={2025}
}

@article{Yin_2024,
  title={A survey on multimodal large language models},
  author={Yin, Shukang and Fu, Chaoyou and Zhao, Sirui and Li, Ke and Sun, Xing and Xu, Tong and Chen, Enhong},
  journal={National Science Review},
  volume={11},
  number={12},
  pages={nwae403},
  year={2024},
  publisher={Oxford University Press}
}

@article{zhang2024mmllmsrecentadvancesmultimodal,
  title={Mm-llms: Recent advances in multimodal large language models},
  author={Zhang, Duzhen and Yu, Yahan and Dong, Jiahua and Li, Chenxing and Su, Dan and Chu, Chenhui and Yu, Dong},
  journal={Findings of the Association for Computational Linguistics: ACL 2024},
  pages={12401--12430},
  year={2024}
}

@article{caffagni2024revolutionmultimodallargelanguage,
  title={The revolution of multimodal large language models: A survey},
  author={Caffagni, Davide and Cocchi, Federico and Barsellotti, Luca and Moratelli, Nicholas and Sarto, Sara and Baraldi, Lorenzo and Cornia, Marcella and Cucchiara, Rita},
  journal={Findings of the association for computational linguistics: ACL 2024},
  pages={13590--13618},
  year={2024}
}

@article{candeep,
  title={Can deep learning beat numerical weather prediction?},
  author={Schultz, Martin G and Betancourt, Clara and Gong, Bing and Kleinert, Felix and Langguth, Michael and Leufen, Lukas Hubert and Mozaffari, Amirpasha and Stadtler, Scarlet},
  journal={Philosophical Transactions of the Royal Society A: Mathematical, Physical and Engineering Sciences},
  volume={379},
  number={2194},
  year={2021},
  publisher={The Royal Society}
}

@inproceedings{
climODE,
title={Clim{ODE}: Climate and Weather Forecasting with Physics-informed Neural {ODE}s},
author={Yogesh Verma and Markus Heinonen and Vikas Garg},
booktitle={The Twelfth International Conference on Learning Representations},
year={2024},
url={https://openreview.net/forum?id=xuY33XhEGR}
}

@article{Interpretable,
  title={Interpretable physics-informed graph neural networks for flood forecasting},
  author={Taghizadeh, Mehdi and Zandsalimi, Zanko and Nabian, Mohammad Amin and Shafiee-Jood, Majid and Alemazkoor, Negin},
  journal={Computer-Aided Civil and Infrastructure Engineering},
  volume={40},
  number={18},
  pages={2629--2649},
  year={2025},
  publisher={Wiley Online Library}
}

@article{Enforcing,
  title={Enforcing analytic constraints in neural networks emulating physical systems},
  author={Beucler, Tom and Pritchard, Michael and Rasp, Stephan and Ott, Jordan and Baldi, Pierre and Gentine, Pierre},
  journal={Physical review letters},
  volume={126},
  number={9},
  pages={098302},
  year={2021},
  publisher={APS}
}

@article{NeuralNetworks,
  title={Use of neural networks for stable, accurate and physically consistent parameterization of subgrid atmospheric processes with good performance at reduced precision},
  author={Yuval, Janni and O'Gorman, Paul A and Hill, Chris N},
  journal={Geophysical Research Letters},
  volume={48},
  number={6},
  pages={e2020GL091363},
  year={2021},
  publisher={Wiley Online Library}
}

@article{GlobalForecasting,
  title={Global forecasting of tropical cyclone intensity using neural weather models},
  author={Gomez, Milton and Berne, Alexis and Beucler, Tom and others},
  journal={arXiv preprint arXiv:2508.17903},
  year={2025}
}

@article{Neuralgeneral,
  title={Neural general circulation models for weather and climate},
  author={Kochkov, Dmitrii and Yuval, Janni and Langmore, Ian and Norgaard, Peter and Smith, Jamie and Mooers, Griffin and Kl{\"o}wer, Milan and Lottes, James and Rasp, Stephan and D{\"u}ben, Peter and others},
  journal={Nature},
  volume={632},
  number={8027},
  pages={1060--1066},
  year={2024},
  publisher={Nature Publishing Group UK London}
}

@article{Physics-informed,
  title={Physics-informed Mamba network for ultra-short-term photovoltaic power forecasting: integrating WGAN-GP augmentation and CEEMDAN-SST decomposition},
  author={Li, Yanmei and Zhang, Yi and Yin, Minghao},
  journal={Renewable Energy},
  pages={124851},
  year={2025},
  publisher={Elsevier}
}

@article{chen2024longvilascalinglongcontextvisual,
  title={Longvila: Scaling long-context visual language models for long videos},
  author={Chen, Yukang and Xue, Fuzhao and Li, Dacheng and Hu, Qinghao and Zhu, Ligeng and Li, Xiuyu and Fang, Yunhao and Tang, Haotian and Yang, Shang and Liu, Zhijian and others},
  journal={arXiv preprint arXiv:2408.10188},
  year={2024}
}

@article{wang2024qwen2vlenhancingvisionlanguagemodels,
  title={Qwen2-vl: Enhancing vision-language model's perception of the world at any resolution},
  author={Wang, Peng and Bai, Shuai and Tan, Sinan and Wang, Shijie and Fan, Zhihao and Bai, Jinze and Chen, Keqin and Liu, Xuejing and Wang, Jialin and Ge, Wenbin and others},
  journal={arXiv preprint arXiv:2409.12191},
  year={2024}
}

@article{dai2026needrealanomalymllm,
  title={No Need For Real Anomaly: MLLM Empowered Zero-Shot Video Anomaly Detection},
  author={Dai, Zunkai and Li, Ke and Liu, Jiajia and Yang, Jie and Qiao, Yuanyuan},
  journal={arXiv preprint arXiv:2602.19248},
  year={2026}
}

@inproceedings{chen2024internvlscalingvisionfoundation,
  title={Internvl: Scaling up vision foundation models and aligning for generic visual-linguistic tasks},
  author={Chen, Zhe and Wu, Jiannan and Wang, Wenhai and Su, Weijie and Chen, Guo and Xing, Sen and Zhong, Muyan and Zhang, Qinglong and Zhu, Xizhou and Lu, Lewei and others},
  booktitle={Proceedings of the IEEE/CVF conference on computer vision and pattern recognition},
  pages={24185--24198},
  year={2024}
}

@inproceedings{li2025vidhallucevaluatingtemporalhallucinations,
  title={Vidhalluc: Evaluating temporal hallucinations in multimodal large language models for video understanding},
  author={Li, Chaoyu and Im, Eun Woo and Fazli, Pooyan},
  booktitle={Proceedings of the IEEE/CVF Conference on Computer Vision and Pattern Recognition},
  pages={13723--13733},
  year={2025}
}

@article{tong2024cambrian1fullyopenvisioncentric,
  title={Cambrian-1: A fully open, vision-centric exploration of multimodal llms},
  author={Tong, Peter and Brown, Ellis and Wu, Penghao and Woo, Sanghyun and Iyer, Adithya Jairam Vedagiri and Akula, Sai Charitha and Yang, Shusheng and Yang, Jihan and Middepogu, Manoj and Wang, Ziteng and others},
  journal={Advances in Neural Information Processing Systems},
  volume={37},
  pages={87310--87356},
  year={2024}
}

@inproceedings{xiao2023florence2advancingunifiedrepresentation,
  title={Florence-2: Advancing a unified representation for a variety of vision tasks},
  author={Xiao, Bin and Wu, Haiping and Xu, Weijian and Dai, Xiyang and Hu, Houdong and Lu, Yumao and Zeng, Michael and Liu, Ce and Yuan, Lu},
  booktitle={Proceedings of the IEEE/CVF Conference on Computer Vision and Pattern Recognition},
  pages={4818--4829},
  year={2024}
}

@inproceedings{11093955,
  title={Rod-mllm: Towards more reliable object detection in multimodal large language models},
  author={Yin, Heng and Ren, Yuqiang and Yan, Ke and Ding, Shouhong and Hao, Yongtao},
  booktitle={Proceedings of the Computer Vision and Pattern Recognition Conference},
  pages={14358--14368},
  year={2025}
}

@article{machinelearning,
  title={Physics-informed machine learning: case studies for weather and climate modelling},
  author={Kashinath, Karthik and Mustafa, Mustafa and Albert, Adrian and Wu, Jean-Luc and Jiang, C and Esmaeilzadeh, Soheil and Azizzadenesheli, Kamyar and Wang, R and Chattopadhyay, Ashesh and Singh, Aakanksha and others},
  journal={Philosophical Transactions of the Royal Society A: Mathematical, Physical and Engineering Sciences},
  volume={379},
  number={2194},
  year={2021},
  publisher={The Royal Society}
}

@article{PaLM-E,
  title={Palm-e: An embodied multimodal language model},
  author={Driess, Danny and Xia, Fei and Sajjadi, Mehdi SM and Lynch, Corey and Chowdhery, Aakanksha and Ichter, Brian and Wahid, Ayzaan and Tompson, Jonathan and Vuong, Quan and Yu, Tianhe and others},
  journal={arXiv preprint arXiv:2303.03378},
  year={2023}
}

@article{VarteX,
  title={VarteX: Enhancing Weather Forecast through Distributed Variable Representation},
  author={Ueyama, Ayumu and Kawamoto, Kazuhiko and Kera, Hiroshi},
  journal={arXiv preprint arXiv:2406.19615},
  year={2024}
}

@article{Learningtoforecast,
  title={Learning to forecast diagnostic parameters using pre-trained weather embedding},
  author={Mitra, Peetak P and Ramavajjala, Vivek},
  journal={arXiv preprint arXiv:2312.00290},
  year={2023}
}

@article{Triformer,
  title={Triformer: Triangular, Variable-Specific Attentions for Long Sequence Multivariate Time Series Forecasting--Full Version},
  author={Cirstea, Razvan-Gabriel and Guo, Chenjuan and Yang, Bin and Kieu, Tung and Dong, Xuanyi and Pan, Shirui},
  journal={arXiv preprint arXiv:2204.13767},
  year={2022}
}

@article{qwen3technicalreport,
  title={Qwen3 technical report},
  author={Yang, An and Li, Anfeng and Yang, Baosong and Zhang, Beichen and Hui, Binyuan and Zheng, Bo and Yu, Bowen and Gao, Chang and Huang, Chengen and Lv, Chenxu and others},
  journal={arXiv preprint arXiv:2505.09388},
  year={2025}
}

@article{wu2024deepseekvl2mixtureofexpertsvisionlanguagemodels,
  title={Deepseek-vl2: Mixture-of-experts vision-language models for advanced multimodal understanding},
  author={Wu, Zhiyu and Chen, Xiaokang and Pan, Zizheng and Liu, Xingchao and Liu, Wen and Dai, Damai and Gao, Huazuo and Ma, Yiyang and Wu, Chengyue and Wang, Bingxuan and others},
  journal={arXiv preprint arXiv:2412.10302},
  year={2024}
}

@article{hunyuanocr2025,
  title={Hunyuanocr technical report},
  author={Team, Hunyuan Vision and Lyu, Pengyuan and Wan, Xingyu and Li, Gengluo and Peng, Shangpin and Wang, Weinong and Wu, Liang and Shen, Huawen and Zhou, Yu and Tang, Canhui and others},
  journal={arXiv preprint arXiv:2511.19575},
  year={2025}
}

@article{wang2025internvl3_5,
  title={InternVL3.5: Advancing Open-Source Multimodal Models in Versatility, Reasoning, and Efficiency},
  author={Wang, Weiyun and Gao, Zhangwei and Gu, Lixin and Pu, Hengjun and Cui, Long and Wei, Xingguang and Liu, Zhaoyang and Jing, Linglin and Ye, Shenglong and Shao, Jie and others},
  journal={arXiv preprint arXiv:2508.18265},
  year={2025}
}

@article{quietrevolution,
  title={The quiet revolution of numerical weather prediction},
  author={Bauer, Peter and Thorpe, Alan and Brunet, Gilbert},
  journal={Nature},
  volume={525},
  number={7567},
  pages={47--55},
  year={2015},
  publisher={Nature Publishing Group UK London}
}

@article{Bauer2021,
  title={The digital revolution of Earth-system science},
  author={Bauer, Peter and Dueben, Peter D and Hoefler, Torsten and Quintino, Tiago and Schulthess, Thomas C and Wedi, Nils P},
  journal={Nature Computational Science},
  volume={1},
  number={2},
  pages={104--113},
  year={2021},
  publisher={Nature Publishing Group US New York}
}

@inproceedings{
ramsauer2020hopfield,
title={Hopfield Networks is All You Need},
author={Hubert Ramsauer and Bernhard Sch{\"a}fl and Johannes Lehner and Philipp Seidl and Michael Widrich and Lukas Gruber and Markus Holzleitner and Thomas Adler and David Kreil and Michael K Kopp and G{\"u}nter Klambauer and Johannes Brandstetter and Sepp Hochreiter},
booktitle={International Conference on Learning Representations},
year={2021},
url={https://openreview.net/forum?id=tL89RnzIiCd}
}

@inproceedings{Cait,
  title={Going deeper with image transformers},
  author={Touvron, Hugo and Cord, Matthieu and Sablayrolles, Alexandre and Synnaeve, Gabriel and J{\'e}gou, Herv{\'e}},
  booktitle={Proceedings of the IEEE/CVF international conference on computer vision},
  pages={32--42},
  year={2021}
}

@article{VIT,
  title={An image is worth 16x16 words: Transformers for image recognition at scale},
  author={Dosovitskiy, Alexey and Beyer, Lucas and Kolesnikov, Alexander and Weissenborn, Dirk and Zhai, Xiaohua and Unterthiner, Thomas and Dehghani, Mostafa and Minderer, Matthias and Heigold, Georg and Gelly, Sylvain and others},
  journal={arXiv preprint arXiv:2010.11929},
  year={2020}
}

@inproceedings{li2025mccd,
  title={Mccd: Multi-agent collaboration-based compositional diffusion for complex text-to-image generation},
  author={Li, Mingcheng and Hou, Xiaolu and Liu, Ziyang and Yang, Dingkang and Qian, Ziyun and Chen, Jiawei and Wei, Jinjie and Jiang, Yue and Xu, Qingyao and Zhang, Lihua},
  booktitle={Proceedings of the Computer Vision and Pattern Recognition Conference},
  pages={13263--13272},
  year={2025}
}

@inproceedings{yang2025nader,
  title={Nader: Neural architecture design via multi-agent collaboration},
  author={Yang, Zekang and Zeng, Wang and Jin, Sheng and Qian, Chen and Luo, Ping and Liu, Wentao},
  booktitle={Proceedings of the Computer Vision and Pattern Recognition Conference},
  pages={4452--4461},
  year={2025}
}

@inproceedings{gupta2023visual,
  title={Visual programming: Compositional visual reasoning without training},
  author={Gupta, Tanmay and Kembhavi, Aniruddha},
  booktitle={Proceedings of the IEEE/CVF conference on computer vision and pattern recognition},
  pages={14953--14962},
  year={2023}
}

@inproceedings{ijcai2025p885,
  title     = {Physics-Assisted and Topology-Informed Deep Learning for Weather Prediction},
  author    = {Zheng, Jiaqi and Ling, Qing and Feng, Yerong},
  booktitle = {Proceedings of the Thirty-Fourth International Joint Conference on
               Artificial Intelligence, {IJCAI-25}},
  publisher = {International Joint Conferences on Artificial Intelligence Organization},
  editor    = {James Kwok},
  pages     = {7958--7966},
  year      = {2025},
  month     = {8},
  note      = {Main Track},
  doi       = {10.24963/ijcai.2025/885},
  url       = {https://doi.org/10.24963/ijcai.2025/885},
}

@inproceedings{
linander2025pear,
title={{PEAR}: Equal Area Weather Forecasting on the Sphere},
author={Hampus Linander and Christoffer Petersson and Daniel Persson and Jan E Gerken},
booktitle={NeurIPS 2025 AI for Science Workshop},
year={2025},
url={https://openreview.net/forum?id=MlnM8lvFSq}
}

@inproceedings{
he2025radarqa,
title={Radar{QA}: Multi-modal Quality Analysis of Weather Radar Forecasts},
author={Xuming He and Zhiyuan You and Junchao Gong and Couhua Liu and Xiaoyu Yue and Peiqin Zhuang and Wenlong Zhang and LEI BAI},
booktitle={The Thirty-ninth Annual Conference on Neural Information Processing Systems},
year={2025},
url={https://openreview.net/forum?id=WlrmpjocNe}
}

@inproceedings{
chen2025agentcaster,
title={AgentCaster: Reasoning-Guided Tornado Forecasting},
author={Michael Chen},
booktitle={NeurIPS 2025 Workshop on Evaluating the Evolving LLM Lifecycle: Benchmarks, Emergent Abilities, and Scaling},
year={2025},
url={https://openreview.net/forum?id=ZTQNYngIyF}
}

@inproceedings{
varambally2026zephyrus,
title={Zephyrus: An Agentic Framework for Weather Science},
author={Sumanth Varambally and Marshall Fisher and Jas Thakker and Yiwei Chen and Zhirui Xia and Yasaman Jafari and Ruijia Niu and Manas Jain and Veeramakali Vignesh Manivannan and Zachary Novack and Luyu Han and Srikar Eranky and Salva R{\"u}hling Cachay and Taylor Berg-Kirkpatrick and Duncan Watson-Parris and Yian Ma and Rose Yu},
booktitle={The Fourteenth International Conference on Learning Representations},
year={2026},
url={https://openreview.net/forum?id=aVeaNahsID}
}

@inproceedings{dhuliawala2024chain,
  title={Chain-of-verification reduces hallucination in large language models},
  author={Dhuliawala, Shehzaad and Komeili, Mojtaba and Xu, Jing and Raileanu, Roberta and Li, Xian and Celikyilmaz, Asli and Weston, Jason},
  booktitle={Findings of the association for computational linguistics: ACL 2024},
  pages={3563--3578},
  year={2024}
}

@inproceedings{gao2025interleaved,
  title={Interleaved-modal chain-of-thought},
  author={Gao, Jun and Li, Yongqi and Cao, Ziqiang and Li, Wenjie},
  booktitle={Proceedings of the Computer Vision and Pattern Recognition Conference},
  pages={19520--19529},
  year={2025}
}

@inproceedings{chen2025egoagent,
  title={EgoAgent: a joint predictive agent model in egocentric worlds},
  author={Chen, Lu and Wang, Yizhou and Tang, Shixiang and Ma, Qianhong and He, Tong and Ouyang, Wanli and Zhou, Xiaowei and Bao, Hujun and Peng, Sida},
  booktitle={Proceedings of the IEEE/CVF International Conference on Computer Vision},
  pages={6970--6980},
  year={2025}
}

@inproceedings{shi2025enhancing,
  title={Enhancing video-llm reasoning via agent-of-thoughts distillation},
  author={Shi, Yudi and Di, Shangzhe and Chen, Qirui and Xie, Weidi},
  booktitle={Proceedings of the Computer Vision and Pattern Recognition Conference},
  pages={8523--8533},
  year={2025}
}

@inproceedings{wang2024cog,
  title={Cog-dqa: Chain-of-guiding learning with large language models for diagram question answering},
  author={Wang, Shaowei and Zhang, Lingling and Zhu, Longji and Qin, Tao and Yap, Kim-Hui and Zhang, Xinyu and Liu, Jun},
  booktitle={Proceedings of the IEEE/CVF Conference on Computer Vision and Pattern Recognition},
  pages={13969--13979},
  year={2024}
}

@inproceedings{yang2025vca,
  title={Vca: Video curious agent for long video understanding},
  author={Yang, Zeyuan and Chen, Delin and Yu, Xueyang and Shen, Maohao and Gan, Chuang},
  booktitle={Proceedings of the IEEE/CVF International Conference on Computer Vision},
  pages={20168--20179},
  year={2025}
}

@inproceedings{ghezloo2025pathfinder,
  title={Pathfinder: A multi-modal multi-agent system for medical diagnostic decision-making applied to histopathology},
  author={Ghezloo, Fatemeh and Seyfioglu, Mehmet Saygin and Soraki, Rustin and Ikezogwo, Wisdom O and Li, Beibin and Vivekanandan, Tejoram and Elmore, Joann G and Krishna, Ranjay and Shapiro, Linda},
  booktitle={Proceedings of the IEEE/CVF International Conference on Computer Vision},
  pages={23431--23441},
  year={2025}
}

@inproceedings{chen2025less,
  title={Less is more: Empowering gui agent with context-aware simplification},
  author={Chen, Gongwei and Zhou, Xurui and Shao, Rui and Lyu, Yibo and Zhou, Kaiwen and Wang, Shuai and Li, Wentao and Li, Yinchuan and Qi, Zhongang and Nie, Liqiang},
  booktitle={Proceedings of the IEEE/CVF International Conference on Computer Vision},
  pages={5901--5911},
  year={2025}
}

@article{liao2025motionagent,
  title={Motionagent: fine-grained controllable video generation via motion field agent},
  author={Liao, Xinyao and Zeng, Xianfang and Wang, Liao and Yu, Gang and Lin, Guosheng and Zhang, Chi},
  journal={arXiv preprint arXiv:2502.03207},
  year={2025}
}

@inproceedings{chen2025t2i,
  title={T2i-copilot: A training-free multi-agent text-to-image system for enhanced prompt interpretation and interactive generation},
  author={Chen, Chieh-Yun and Shi, Min and Zhang, Gong and Shi, Humphrey},
  booktitle={Proceedings of the IEEE/CVF International Conference on Computer Vision},
  pages={19396--19405},
  year={2025}
}

@inproceedings{yao2024promptcot,
  title={Promptcot: Align prompt distribution via adapted chain-of-thought},
  author={Yao, Junyi and Liu, Yijiang and Dong, Zhen and Guo, Mingfei and Hu, Helan and Keutzer, Kurt and Du, Li and Zhou, Daquan and Zhang, Shanghang},
  booktitle={Proceedings of the IEEE/CVF conference on computer vision and pattern recognition},
  pages={7027--7037},
  year={2024}
}
\end{document}